\documentclass[acmtog]{acmart}
\AtBeginDocument{%
  }

\usepackage{graphics}
\usepackage{color}
\usepackage{multirow}
\usepackage{makecell}
\usepackage{soul} %
\usepackage{pifont}%
\usepackage{tabularx}
\usepackage{svg}
\usepackage{amsmath}
\usepackage{subcaption}
\usepackage{graphicx}
\usepackage{booktabs}
\usepackage{cleveref}
\definecolor{turquoise}{cmyk}{0.65,0,0.1,0.3}
\definecolor{purple}{rgb}{0.65,0,0.65}
\definecolor{dark_green}{rgb}{0, 0.5, 0}
\definecolor{orange}{rgb}{0.8, 0.6, 0.2}
\definecolor{red}{rgb}{0.8, 0.2, 0.2}
\definecolor{darkred}{rgb}{0.6, 0.1, 0.05}
\definecolor{blueish}{rgb}{0.0, 0.3, .6}
\definecolor{light_gray}{rgb}{0.7, 0.7, .7}
\definecolor{pink}{rgb}{1, 0, 1}
\definecolor{greyblue}{rgb}{0.25, 0.25, 1}
\definecolor{forestgreen}{rgb}{0.0, 0.2, 0.13}
\definecolor{darkolivegreen}{rgb}{0.33, 0.42, 0.18}

\newcommand{\moniker}{GenLit}

\newcommand{\datasetname}{\textsc{Objaverse-GenLit}}

\newcommand{\datasetReal}{Real Data}
\newcommand{\datasetRealSingle}{Real\-Data-\-\textsc{Single\-Object}}
\newcommand{\datasetRealMulti}{Real\-Data-\textsc{Multi\-Object}}
\newcommand{\eg}{e.g. }

\newcommand{\singleobjecttrain}{\textsc{Single\-Object}}
\newcommand{\multiobjecttrain}{\textsc{Multi\-Object-\-Flying\-Light}}
\newcommand{\singleobjecttest}{Single\-Object-\-Test}

\crefname{figure}{Fig.}{Figs.}
\Crefname{figure}{Fig.}{Figs.}

\crefname{table}{Tab.}{Tabs.}
\Crefname{table}{Tab.}{Tabs.}

\crefname{section}{Sec.}{Secs.}
\Crefname{section}{Sec.}{Secs.}

\crefname{subsection}{Sec.}{Secs.}
\Crefname{subsection}{Sec.}{Secs.}

\copyrightyear{2025}
\acmYear{2025}
\acmConference[SA Conference Papers '25]{SIGGRAPH Asia 2025 Conference Papers}{December 15--18, 2025}{Hong Kong, Hong Kong}
\acmBooktitle{SIGGRAPH Asia 2025 Conference Papers (SA Conference Papers '25), December 15--18, 2025, Hong Kong, Hong Kong}\acmDOI{10.1145/3757377.3763970}
\acmISBN{979-8-4007-2137-3/2025/12}

\begin{document}

\title{GenLit: Reformulating Single-Image Relighting as Video Generation}

\author{Shrisha Bharadwaj}
\orcid{0009-0000-7390-456X}
\authornote{Equal contribution, listed alphebetically.}
\authornote{corresponding author.}
\affiliation{%
 \institution{Max Planck Institute for Intelligent Systems}
 \city{Tübingen}
 \country{Germany}}
\email{shrisha.bharadwaj@tuebingen.mpg.de}
\author{Haiwen Feng}
\orcid{0009-0005-7629-0547}
\authornotemark[1]
\affiliation{%
 \institution{Max Planck Institute for Intelligent Systems}
 \city{Tübingen}
 \country{Germany}}
\email{haiwen.feng@tuebingen.mpg.de}
\author{Giorgio Becherini}
\orcid{0009-0005-8770-8144}
\affiliation{%
 \institution{Max Planck Institute for Intelligent Systems}
 \city{Tübingen}
 \country{Germany}}
\email{giorgio.becherini@tuebingen.mpg.de}
\author{Victoria Fernandez Abrevaya}
\orcid{0000-0002-9829-4929}
\affiliation{%
 \institution{Max Planck Institute for Intelligent Systems}
 \city{Tübingen}
 \country{Germany}}
\email{victoria.abrevaya@tuebingen.mpg.de}
\author{Michael J.~Black}
\orcid{0000-0001-6077-4540}
\affiliation{%
 \institution{Max Planck Institute for Intelligent Systems}
 \city{Tübingen}
 \country{Germany}}
\email{black@tuebingen.mpg.de}

\begin{abstract}
Manipulating the illumination of a 3D scene within a single image represents a fundamental challenge in computer vision and graphics. This problem has traditionally been addressed using inverse rendering techniques, which involve
explicit 3D asset reconstruction and costly ray-tracing simulations. Meanwhile, recent advancements in visual foundation models suggest that a new paradigm could soon be possible -- one that replaces explicit physical models with networks that are trained on large amounts of image and video data. In this paper, we exploit the implicit scene understanding of a video diffusion model, particularly Stable Video Diffusion, to relight a single image. 
We introduce \moniker{}, a framework that distills the ability of a graphics engine to perform light manipulation into a video-generation model, enabling users to directly insert and manipulate a point light in the 3D world within a given image and generate results directly as a video sequence. 
We find that a model fine-tuned on only a small synthetic dataset generalizes to real-world scenes, enabling single-image relighting with plausible and convincing shadows and inter-reflections. Our results highlight the ability of video foundation models to capture rich information about lighting, material, and shape, and our findings indicate that such models, with minimal training, can be used to perform relighting without explicit asset reconstruction or ray-tracing. 
\end{abstract}




\begin{teaserfigure}
\centering
\includegraphics[width=1.0\textwidth]{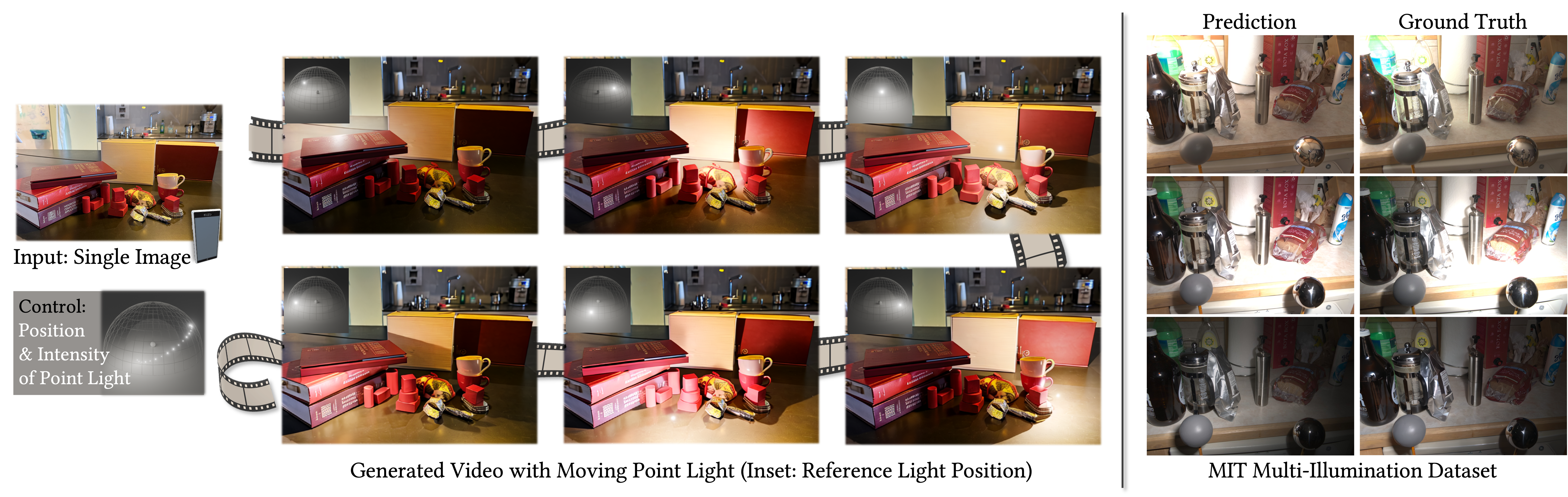}
  \caption{\textbf{GenLit} is a new framework for single-image relighting. We reformulate the task as image-to-video generation, keeping the scene and the object in the scene static while generating the lighting changes by controlling the position of the light source. Our method is trained on a synthetic dataset and  generalizes to real images. Left: \moniker{} takes an image (here, captured with a phone) along with the control signal (position and intensity of light source) and inserts a point light that can move arbitrarily around the scene along a desired trajectory, relighting the scene. The relit results have direct lighting effects like plausible shape-appropriate shadows and indirect effects such as diffuse interreflections. Right: Qualitative comparisons on the testset of MIT Multi.\ Illumn \cite{murmann19} dataset where the relit results closely resemble the ground truth.
}\label{fig:teaser}
\end{teaserfigure}

\maketitle

\section{Introduction}
From digital content creation to augmented reality, modifying the lighting of a scene in a single image enables a wide range of creative and interactive applications.
A substantial body of research has addressed the problem of modifying \emph{global} scene illumination for indoor/outdoor scenes~\citep{Li2022PhysicallyBasedEO, kocsis2024lightit, griffiths2022outcast, yu2020self} and objects~\citep{pandey2021total, papantoniou2023relightify, ponglertnapakorn2023difareli}. 
However, the task of \emph{near-field} relighting -- the adding of local light sources that may even be visible in the scene -- has received much less attention. 
Local illumination offers complementary strengths to global relighting: it reveals high-frequency surface details and scene structure through cast shadows, specular reflections, and shading effects; these have practical relevance for applications such as virtual product showcasing and post-production editing.

Single-image relighting is inherently complex, as it is an ill-posed problem due to multiple ambiguities relating to the source light and albedo, lack of depth information, and 
unknown materials.
A common solution involves using an \emph{inverse-rendering} framework~\citep{jakob2022mitsuba3, barron2014shape, li2020inverse},
whereby explicit 3D assets for physically based rendering (PBR) are recovered (geometry, material properties, and scene illumination), and re-rendered under specified lighting conditions. However, effective simulation of lighting requires not only a precise estimation of these assets, but also an accurate reproduction of the behaviour of light with respect to individual materials, casting direct shadows and bouncing of light rays. %
Explicitly recovering all of this is difficult, prone to errors, and can be computationally expensive.

Recent progress in image and video foundation models~\citep{rombach2021highresolution, blattmann2023videoldm} suggests that a new paradigm may be possible, where explicit PBR asset reconstruction might be replaced by a learned image encoder and conditioning module, and the simulation of complex ray tracing interactions are replaced by the generative model itself. The extensive training data behind such models enable them to learn and recreate real-world images at a pixel level, eliminating the need for {\em explicit} physics-based representations.
Video diffusion models, in particular, have shown evidence of 3D scene understanding~\citep{blattmann2023svd, voleti2024sv3d} and can generate temporally consistent video that maintains the invariance of scene attributes, given a single image as input. 
This suggests that such models may have a significant understanding of the world and be able to reason about material properties as well as light--matter interactions. 

Our solution is based on the observation that modifying the lighting in a scene involves change. Video diffusion models are trained to model change over time, including changes in illumination. Based on this observation, our key insight is that we can {\em leverage video diffusion models to solve the single-image relighting problem}, effectively exploiting their implicit 3D representation of the world.
 
We focus on near-field relighting with a point light source to obtain fine-grained and flexible control over local illumination. We call our method \textbf{\moniker{}}, which performs generative relighting by enhancing
an image-to-video model (Stable Video Diffusion (SVD)~\citep{blattmann2023svd}) with an interpretable control signal~\citep{zhang2023controlnet}. Specifically, \moniker{} takes a single image and generates a video of a moving light that is controlled by a 5D lighting vector, which includes the point light's 3D location and intensity and the ambient light's intensity.
We train our method on a curated synthetic dataset consisting of 1436 objects and show generalization to real-world scenes with notable complexity (see \cref{fig:teaser}). We model the background and foreground holistically without segmenting the objects to ensure that the shadows of the foreground object are cast on the surface (or background) or on other objects. We observe that these shadows are \textit{crucial} for realism, and this is validated by our evaluations. %

We evaluate on two broad settings, \singleobjecttrain{} and \multiobjecttrain{} and show generalization to real images in both settings.  For \multiobjecttrain{}, the motion of the point light is arbitrary and freely moves around, above, and even between the objects in the scene. Thus, the motions are unconstrained, allowing the light to fly through the scene. 
This flying light may be outside the field of view or visible in the scene; the motion of this light gives a strong perceptual sense of the 3D scene without ever explicitly reconstructing it. 
Additionally, we conduct quantitative evaluations on real images by evaluating on the MIT Multi-Illumination Dataset~\citep{murmann19}. Compared to state-of-the-art single-image relighting methods~\citep{Yi_2023_CVPR, iclight, Jin2024NEURIPS_Neural_Gaffer_Relighting, zeng2024dilightnet, zhang2024latentintrinsicsemergetraining, DiffusionRenderer, Xing2024luminet}, we obtain significantly greater accuracy on both synthetic and real datasets. 

In summary, we introduce a framework that distills the ability to relight a single image with a point light from
data rendered by a graphics engine to a pre-trained image-to-video generative model, and the finetuned system generalizes to in-the-wild images. Our work reveals that video foundation models have a sufficiently rich understanding of light, materials, and shape to support single-image controllable relighting, without the need to explicitly capture and design PBR assets (geometry, material, illumination). Moreover, by inserting the point light between/around the objects in a scene, \moniker{} can synthesize direct effects such as distinct shadows and local indirect effects like diffuse interreflections without expensive ray-tracing. Our results show that \moniker{} can synthesize plausible and convincing shape- and light-appropriate shadows that generalize to \emph{unseen} real-world objects as well as real images of complex scenes with diverse material properties. We believe that \moniker{} is a step-forward in terms of control and flexibility, as the inserted point-light has free motion. Since our method can convincingly generalize to a wide range of in-the-wild objects, it has potential for applications like product showcases, content creation, and interactive visual experiences in cases where it is time-consuming and challenging to manually design 3D assets.
Code, model weights, and the dataset can be found in \url{https://genlit.is.tue.mpg.de/}.

\section{Related Work}
We focus our discussion on methods that manipulate the light from a single image, either via explicit inverse rendering or end-to-end neural methods, and on controllable synthesis with diffusion models.
 
\subsection{Inverse Rendering}
Inverse rendering approaches estimate explicit properties such as geometry, material, and illumination from a single image, which can then be used to re-render the scene under novel lighting conditions. 
Early work employs optimization-based approaches along with priors~\citep{barron2014shape} and a large paired dataset of reflectance images~\citep{bell2014intrinsic}. More recent methods leverage deep neural networks to estimate these for indoor scenes~\citep{sengupta2019neural, li2020inverse, wang2021learning, Li2022PhysicallyBasedEO, kocsis2024iid, luo2024intrinsicdiffusion}, outdoor scenes~\citep{yu2019inverserendernet}, or single-objects~\citep{taniai2018neural}. A few methods use multi-view images to disentangle geometry and lighting, and learn a neural BRDF representation~\citep{srinivasan2021nerv, zhang2021nerfactor}. %
The majority of prior work cannot capture objects with complex, spatially varying BRDFs from only a single image, as they assume diffuse materials or require multi-view images. Methods that take single-images and estimate intrinsic images~\citep{zhu2022learning} use a realistic synthetic dataset to train a convolutional neural network. However, they only perform object insertion and not relighting of the scene. Recent work~\citep{du2023generative, Bhattad2023StyleGANKN, kocsis2024iid} leverages %
generative models to tackle out-of-domain generalization. But, their extracted intrinsic representations are still limited in terms of material properties, or the quality of the estimated properties are insufficient to support good relighting. 
In contrast, our goal is to leverage the implicit 3D representation of image-to-video models to handle complex and out-of-domain, objects.
\begin{figure}
\centering
  \includegraphics[width=1\linewidth]{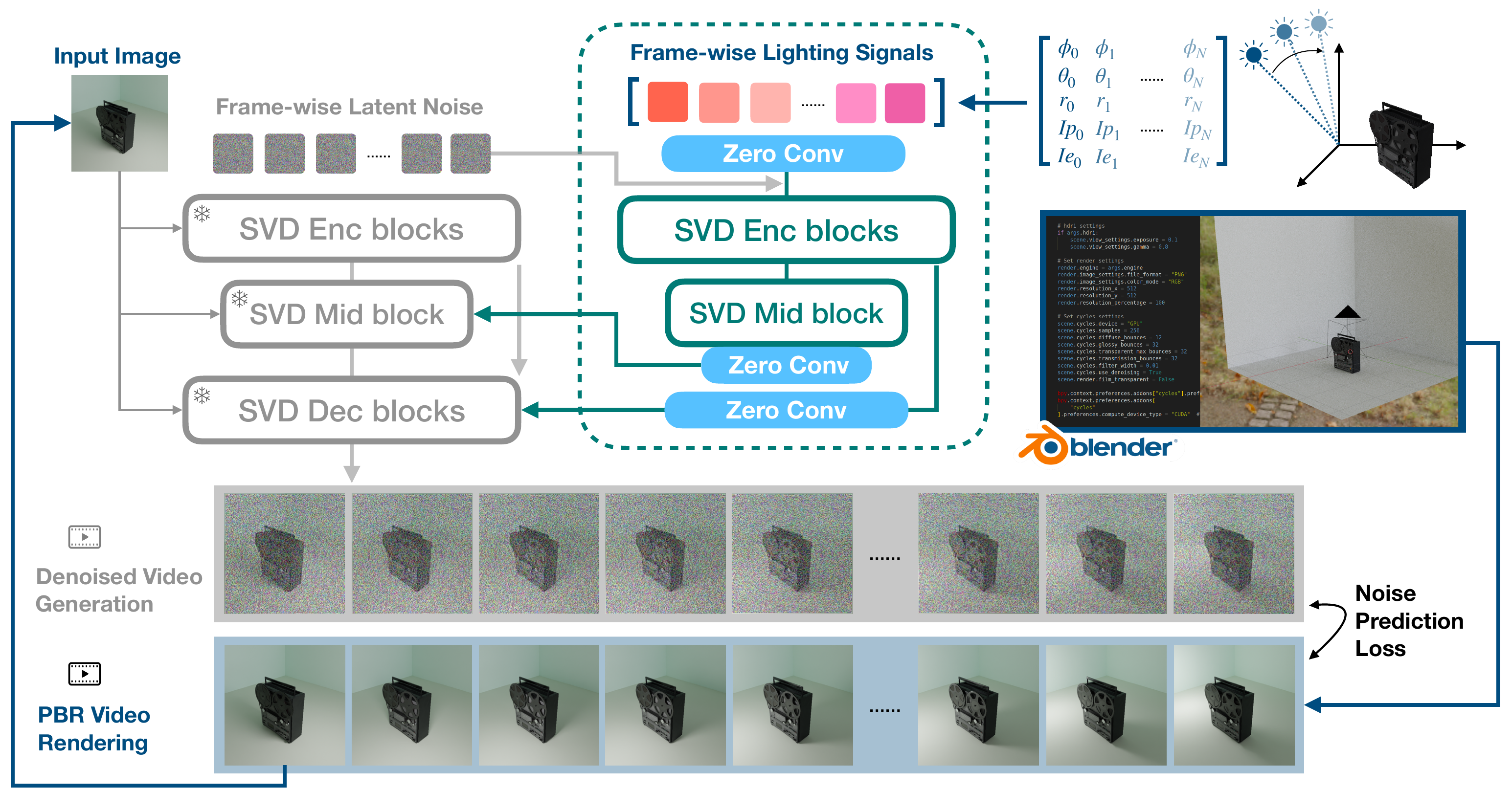}
  \caption{
    \textbf{Overview}. 
    We generate synthetic videos with a point light in motion while the scene is static. The first frame is fed to the generative branch (grey) as conditioning input, while per-frame lighting signals (5D vector) are provided as global information to the control branch (green). 
    }
  \label{fig:pipeline}
\end{figure}

\subsection{Neural Relighting} %
\label{ssec:related_neural}
An alternative to explicit reconstruction is to use implicit neural approaches, either for direct relighting or as priors. Several works employ expensive light-stage setups~\citep{sun2019single, pandey2021total, mei2024holorelighting, rao2022vorf, rao2024lite2relight} to learn data-driven priors with neural networks, 
or use GAN-based models~\citep{deng2023lumigan, ranjan2023facelit} and massive in-the-wild datasets to achieve generalization. However, they are class-specific and cannot be easily extended to arbitrary objects. Li et al.~\shortcite{li2020inverse} use physically-based rendering to edit existing light sources for a given image, however, the approach is limited to the light sources within the scene. With the rise of diffusion models, several relighting approaches have been proposed that leverage diffusion priors, \eg to estimate HDR maps~\citep{lyu2023dpi, Phongthawee2024CVPR_DiffusionLight_Light_Probes}, insert objects~\citep{Liang2024ECCV_Photorealistic_Object_Insertion}, or de-light~\citep{Chen2024ECCV_IntrinsicAnything_Learning_Diffusion}. More closely related to us, \citet{bashkirova_lasagna_2023, kocsis2024lightit, Zeng2024ARXIV_DiLightNet_Fine_grained} aim to control the lighting of an image generated by a text-to-image model. \citet{bashkirova_lasagna_2023} focus on class conditional control and are restricted to 12 discrete lighting directions; \citet{kocsis2024lightit} focus on large outdoor scenes; while\citep{Zeng2024ARXIV_DiLightNet_Fine_grained} and  \citep{Jin2024NEURIPS_Neural_Gaffer_Relighting} employ environment maps as input, which do not generate fine-grained details when illuminated with close light sources. Some works treat the task of relighting as an image-to-image translation task~\citep{iclight, Xing2024luminet}, but, controllability is limited to the contents of the image. \citet{zeng2024rgb} decompose images to intrinsic channels, perform edits, and use the edited intrinsic channels to recompose/relight the image. However, the approach struggles to retain the consistency of the original image and leads to identity shifting. Instead, \moniker{} leverages a video diffusion model, which plausibly has a deeper implicit understanding of changes that occur with light manipulations. Moreover, since \moniker{} is designed to keep the image static and move the light, the model preserves the identity of the scene. Concurrent work, DiffusionRenderer ~\citep{DiffusionRenderer}, extends \citet{zeng2024rgb} with a video model. However their method is still dependent on disentangling intrinsic properties such as materials, lighting, and geometry and is restricted to materials that can be defined with roughness and metalness.

\subsection{Controllable Video Synthesis with Diffusion Models}
\label{ssec:related_video}
With the emergence of methods  for fine-grained control of text-to-image (T2I) models~\citep{zhang2023controlnet, zavadski2023controlnetxs, mou2023t2i}, several extensions have been proposed to handle videos, \eg for body-pose control~\citep{hu2023animateanyone, xu2023magicanimate, chang2024magicpose}, 
camera and object movement control~\citep{wang2023motionctrl}, and last-frame control~\citep{Zeng2023MakePD,xing2023dynamicrafter, Feng2024ExplorativeIO, Jain2024VideoIW}. To the best of our knowledge, \moniker{} is the first work that investigates controllable relighting focusing on local illumination with an image-to-video (I2V) generative model.

\section{Method}
Our goal is to realistically relight a single image with a point light source without performing explicit inverse graphics. To this end, we exploit the implicit scene understanding of an image-to-video diffusion model to  reformulate the problem of single-image relighting as a video generation task, where the scene and objects remain static while illumination changes dynamically over time. We hypothesize that video models implicitly represent physical properties of the world, enabling them to model how light interacts not only with the incident materials, but also with the surrounding environments. Specifically, we fine-tune a controllable variant of Stable Video Diffusion 
(SVD)~\citep{blattmann2023videoldm, blattmann2023svd} using a custom-designed control signal that continuously modifies the position and intensity of a point light source. To enable this, we construct a new dataset of synthetic objects, \datasetname{}, described in \cref{ssec:method_dataset}. 
An overview of our pipeline is shown in \cref{fig:pipeline}. 

\subsection{Background}
\label{ssec:method_svd}
Stable Video Diffusion~\citep{blattmann2023videoldm, blattmann2023svd} is a high-resolution image-to-video model based on Stable Diffusion \cite{rombach2021highresolution}, which introduces temporal blocks that learn to align the generated frames in a temporally consistent manner. Given a single image, SVD generates a video sequence of \(N\) frames, denoted by \(\mathbf{x} = \{x_0, x_1, \ldots, x_{N-1}\}\). This sequence is constructed through a denoising diffusion process where, at each step \(t\), a conditional 3D-UNet, \(\Phi\), is used to iteratively denoise the sequence: $\mathbf{x}^t = \Phi(\mathbf{x}^{t-1}, c)$.
Here, \(c\) represents the conditioning information, which contains the CLIP \cite{Radford2021LearningTV} embedding of the single image input and the latents generated by the Stable Diffusion's VAE for the input image. This conditioning provides a consistent reference of the input image to the UNet and helps in retaining the shape and material through the generation process when the scene light is changed. 

\subsection{Image Relighting as Controllable Video Synthesis} 
\paragraph{Control signal} 
\label{ssec:light_representation}
We use a single point light as the new light source, which is defined by two time-dependent parameters: 3D position and intensity. To remove the effects caused by the ambient light already present in the image, we train our model to first ``dim'' the intensity of the environment, after which we introduce the light motion. 
Specifically, for each generated video, the environment map intensity is gradually reduced in the first four frames until it reaches $20\%$ of its initial value. The intensity of the inserted point light is also gradually increased, from 0 to 120 lumens, starting from the second frame. See the Supplemental Document for a sensitivity analysis of these parameters. From the fifth to the last frame, the point light is moved to different positions, as specified by the control signal. The point light's location at each frame $i$ is defined in polar coordinates $(\phi_i, \theta_i, r_i)$, with $\phi_i \in [0, 360]$, and $\theta_i \in [45^{\circ}, 80^{\circ}]$ including a full range of circular motion. The conditioning signal is a sequence of 5D vectors $[\Vec{l}_1, \dots, \Vec{l}_N]$, where $\Vec{l}_i = [ (\phi_i, \theta_i, r_i, I_{p_i}, I_{e_i});  i=1\dots N ]$,
with $I_{p_i}$ and $I_{e_i}$ scalar values representing the intensity of the point light and of the environment light, respectively.

\paragraph{Architecture} We start from the pretrained SVD model that generates a video from a single image, and extend it with an approach similar to ControlNet \cite{zhang2023controlnet}. We incorporate the control signal by broadcasting each $\Vec{l}_i$ onto an image $L_i \in \mathbb{R}^{H \times W \times 5}$, which is fed to a trainable copy of the U-Net encoder \(\Psi\)  that predicts control conditioning features. These features are added to SVD's encoder and middle block \(\Phi\), and directly influence the generation through the U-Net's decoder. The pre-trained SVD model is frozen and only the weights of the control branch ($\Psi$) are updated. The conditional denoising process is given by: $\mathbf{x}^t = c_{\text{skip}}(\sigma)\,\mathbf{x}^{t-1} \;+\; c_{\text{out}}(\sigma)\,\Phi(\mathbf{x}^{t-1}, c, \Psi(\mathbf{L}))$, where, $\mathbf{L}=\{L_i|i = 1,\dots,N\}$ is the control signal, $\sigma$ is the standard deviation of the Gaussian noise, $c_{\text{skip}}(\sigma)=\tfrac{1}{1+\sigma^2}$ and $c_{\text{out}}(\sigma)=\tfrac{\sigma}{\sqrt{1+\sigma^2}}$ (for $\sigma_{\text{data}}=1$) are scaling coefficients following the EDM Framework \cite{Karras2022edm}. Given clean samples $\mathbf{x}^0$, the model is fine-tuned with the following loss: $\mathcal{L} = \mathbb{E}_{\mathbf{x}^0,\epsilon,\sigma}\Big[ w(\sigma)\,\|\mathbf{x}^t - \mathbf{x}^0\|_2^2 \Big]$,  with weights $w(\sigma)=\tfrac{1+\sigma^2}{\sigma^2}$.

\begin{figure}
    \centering
    \newcommand{\mywidth}{0.45}
\includegraphics[width=\mywidth\textwidth]{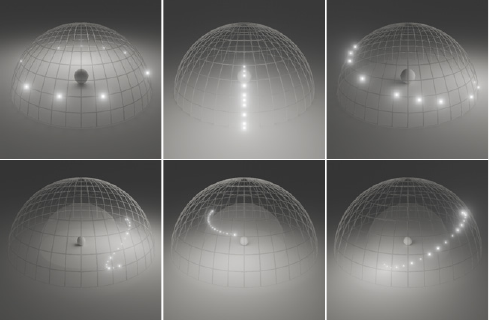}
\caption{Examples of light trajectories for \singleobjecttrain{} (Top) and \multiobjecttrain{} (Bottom)}
\label{fig:light_trajectories}
\end{figure}
\subsection{The \datasetname{} Dataset}
\label{ssec:method_dataset}
To train our model we introduce \datasetname, a dataset of synthetic videos where static objects are illuminated by a moving point light source, capturing diverse light--object interactions that are challenging to model. Each video is rendered using the Cycles path-tracing rendering engine in Blender 3.2.2. We include two different types of scene: (1) \singleobjecttrain{}: where a single object is placed in the center of the scene. (2) \multiobjecttrain{}: where several objects are randomly scattered across the scene. For a multi-object scene, we insert a point-light that flies around the scene. For each video, the scenes are observed from a fixed viewpoint, while a single point light moves around it, dynamically changing the local illumination. We describe \datasetname{} in further detail below.

\paragraph{3D Assets} The objects were sourced from the Objaverse \cite{deitke2024objaverse} dataset and from the Poly Haven \cite{polyhaven} public 3D asset library.
The Objaverse dataset offers a diverse collection of 3D meshes and materials spanning 945 LVIS~\cite{gupta2019lvis} category annotations. This dataset includes not only common categories like cars and chairs but also intricate and unconventional objects that reflect real-world complexity. Please refer to the Supplemental Document for examples. For training, we manually filtered 1200 objects that exhibit high-quality geometry and materials. The Poly Haven 3D asset library includes detailed objects in terms of geometry, shading and texturing. We use a subset of this library composed of 236 objects. In total, we use 1436 3D assets sourced from these two datasets.

\paragraph{Constructing a Scene.} First, we place a single textured ground plane, with textures sourced from Poly Haven. A total of 370 textures are used, excluding classes such as rock, terrain, roofing, and outdoor to prevent uneven surface patterns. The textured plane is then randomly rotated to enhance visual diversity. Furthermore, to create diverse ambient lighting conditions, we randomly select an HDRI map from a curated set of 620 maps provided by Poly Haven. The selection is filtered to exclude maps that contains strong direct sunlight. The camera is positioned overhead, oriented toward the center of the scene. A random perturbation of $\pm 0.3\,\mathrm{m}$ is added to its elevation; since the camera is always looking at the center of the scene, this also results in a change in the pitch angle.

To construct a \singleobjecttrain{} scene, we place an Objaverse object at the center and apply a random rotation around the up-axis to vary the object's orientation relative to the camera viewpoint. Thus, we have 1200 objects for training \singleobjecttrain{}. 
To assemble each \multiobjecttrain{} scene, we sample one object from Poly Haven and up to ten objects from Objaverse. We ensure that at least one Poly Haven object is placed in each scene because of the high-fidelity geometry and materials, and create 1000 random training scenes for \multiobjecttrain{}. In this multi-object setup, each object gets assigned a random rotation along the up-axis and is sequentially placed along a spiral path radiating from the center. To prevent interpenetration, we perform collision checks using the objects' bounding boxes. If a collision is detected, we continue sampling subsequent positions along the spiral until a collision-free placement is found. Please refer to the Supplemental Document for visualizations of both scene types and additional details.

\paragraph{Light}
As described in \cref{ssec:light_representation}, the location of the point light is represented by polar coordinates. To generate diverse light trajectories for the \singleobjecttrain{} scene, we generate a grid of light positions by sampling $M$ evenly spaced $\theta$ and $\phi$ values and creating horizontal, vertical, and diagonal light motions by interpolating between selected points. During training, we set $M = 128$ and randomly sample 40 light motions between pairs of points along the sphere. For testing, we set $M = 64$ to ensure there is no overlap with the train light motions. We fix the radius $r = 1$, the elevation angle to $45^\circ \leq \phi \leq 80^\circ$, and allow the azimuth angle to vary across the full range $0^\circ \leq \theta \leq 360^\circ$. This configuration results in several light trajectories that form loops. For the multi-object setup, we generate Bézier curves and spirals by sampling polar coordinates within the following ranges: $0.8 \leq r \leq 1.5$, $45^\circ \leq \phi \leq 80^\circ$, and $0^\circ \leq \theta \leq 360^\circ$. As illustrated in \cref{fig:light_trajectories}, we randomly sample 40 light trajectories for each scene during training, consisting of 13 Bézier curves, 13 spirals, and 14 hybrid paths that combine characteristics of both Bézier curves and spirals. For testing, we randomly sample 7 of each curve category.

The light trajectories for \singleobjecttrain{} are designed to serve as a systematic testbed to closely observe local illumination effects and conduct thorough evaluations. Whereas, for the \multiobjecttrain{} scene, the motion of the light has more freedom and is relatively unconstrained to aid generalization to scenes with diverse compositions of objects. The motion of the light appears to fly through the scene due to the arbitrary nature of the motions. To incorporate this effect, we use the Blender compositor to apply a glow effect to a mask of the moving light. More details about this post-processing step can be found in the Supplemental Document.

\paragraph{Testing Split} 
1- \singleobjecttest{}: 
\label{data_des:test_single_object}
This test set has 40 objects rendered from two randomly sampled views, and each object has 40 light trajectories, resulting in a total of 3.2k videos and 45k frames. Of these 40 objects, 6 are classic models commonly used in computer graphics (Armadillo, Nefertiti, Stanford Bunny, Teapot, XYZ Dragon, Teapot-2), which we manually verified are not part of the LVIS classification of Objaverse; 20 objects are sampled from Poly Haven by randomly selecting one object per class/category; the remaining 14 objects are sampled from Objaverse by ensuring they belong to different LVIS categories independent of training categories. 

2- \datasetReal{}: 
\label{data_des:test_real}
To evaluate zero-shot generalization to real-world images, we capture photographs of various objects using an iPhone 15. These objects are placed on flat surfaces under ambient illumination. For the \singleobjecttrain{} case we place a single object approximately in the center of the frame and refer to this testset as \datasetRealSingle{}. For the \multiobjecttrain{} setting we randomly place the objects around the scene. Additionally, to specifically observe indirect lighting effects, we place objects inside a box with white colored walls and an infinity cove. We refer to this entire set as \datasetRealMulti{}. Since this test set does not include relighting ground truth, it is intended solely for qualitative evaluation and the perceptual study. 

\section{Evaluation}
First, in \cref{eval:object_level}, we evaluate the relighting quality and compare with state-of-the-art methods for single-image relighting. Next, we qualitatively evaluate \moniker{}'s ability to generalize to real-world samples. We use the model trained on \singleobjecttrain{} for both the evaluations above. Next, in \cref{sec:scene_level}, we conduct qualitative evaluations on \datasetReal{}, along with a perceptual study, using the model trained on \multiobjecttrain{}. Finally, we conduct evaluations on the MIT Multi-Illum.\ Dataset \cite{murmann19} and show that \moniker{} can scale to real-world captures containing multiple objects. The quantitative evaluations are measured using image-space metrics, namely Root Mean Squared Error (RMSE), perceptual loss (LPIPS) \cite{zhang18lpips}, structural similarity index measure (SSIM) \cite{SSIM} and PSNR against the ground truth images. Please refer to the Supplemental Video for results. 

\paragraph{Perceptual Study}
We conducted a perceptual study to evaluate the visual quality of the generalization results. Participants were shown the input image along with the generated video and a video of reference light position. They were asked: \textit{``Judge the realism of the lighting in the predicted video based on how accurately it follows the reference light trajectory  and how well it illuminates the scene in the input image"}. Moreover, they were additionally instructed to consider the plausibility of the shadows with respect to the position of the light source, realism of the generated materials and, finally, whether all the objects were illuminated realistically. They rated each result on a five-point Likert scale. Additionally, the study consisted of catch-trials where obviously incorrect reference trajectories were paired with the generated videos. See the Supplemental Document for details.

\begin{table}[htbp]
\centering
\caption{\textbf{Quantitative Evaluations: Object-Level:} Baseline comparisons for single-image relighting against diffusion-based baselines \cite{iclight, Jin2024NEURIPS_Neural_Gaffer_Relighting, zeng2024dilightnet, DiffusionRenderer} and an inverse-rendering based method \cite{Yi_2023_CVPR} on \singleobjecttest{}. For fairness, the evaluation is performed only on the foreground objects. Bold is best.}
\begin{tabular}{lcccc}
\toprule
         & RMSE $\downarrow$ & LPIPS $\downarrow$ & SSIM $\uparrow$ & PSNR $\uparrow$ \\
\midrule
WS-SIR   & 0.0766 & 0.0530 & 0.9364  & 37.5996 \\
Neural Gaffer & 0.0456 & 0.0596 & 0.9377 & 38.5483 \\
IC-Light & 0.0583 & 0.0450  & 0.9505  & 39.0976 \\
DiLightNet & 0.0764 & 0.0645 & 0.9288 & 38.9696 \\
Diffusion Rend.\ & 0.0453 & 0.0308 & 0.9634 & 39.2117  \\
Ours     & \textbf{0.0309} & \textbf{0.0209} & \textbf{0.9780} & \textbf{40.7174}  \\
\bottomrule
\end{tabular}
\label{tab:baseline_comparison}
\end{table}

\begin{figure*}
    \centering
    \newcommand{\mywidth}{1}
\includegraphics[width=\mywidth\textwidth]{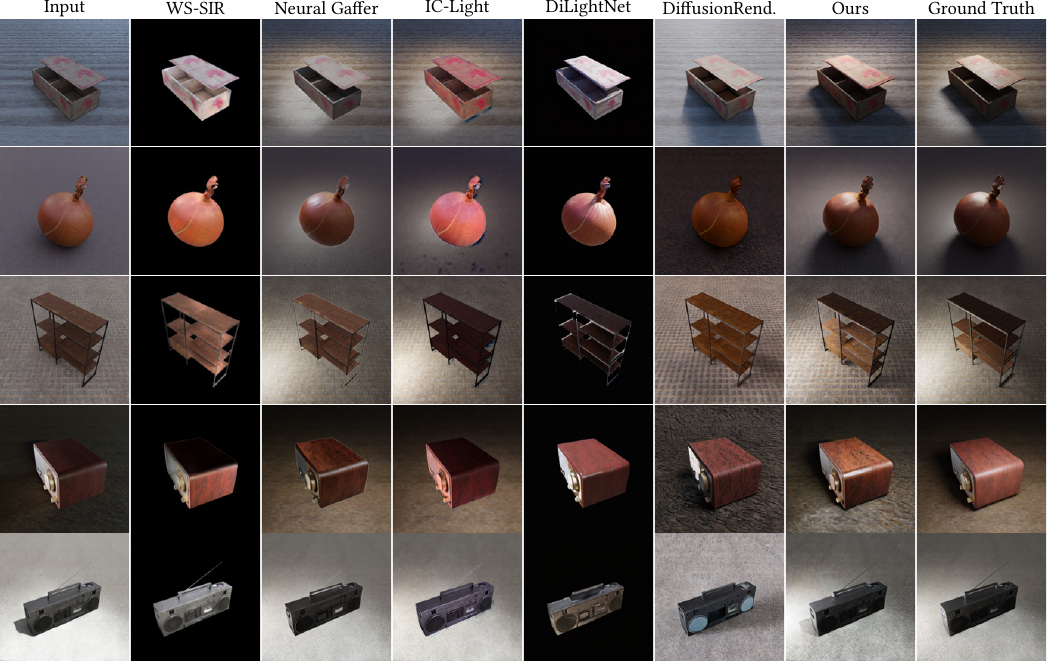}
\caption{\textbf{Object-Level:} \textbf{Qualitative Evaluation:} Baseline comparisons with WS-SIR \cite{Yi_2023_CVPR}, Neural Gaffer \cite{Jin2024NEURIPS_Neural_Gaffer_Relighting}, IC-Light \cite{iclight}, DiLightNet \cite{zeng2024dilightnet} and Diffusion Renderer \cite{DiffusionRenderer}. We provide ground-truth rendered backgrounds for IC-Light \& Neural Gaffer.}
\Description{Baseline Comparisons}
\label{fig:baseline_fig}
\end{figure*}
\begin{figure*}
    \centering
    \newcommand{\mywidth}{1}
    \includegraphics[width=\mywidth\textwidth]{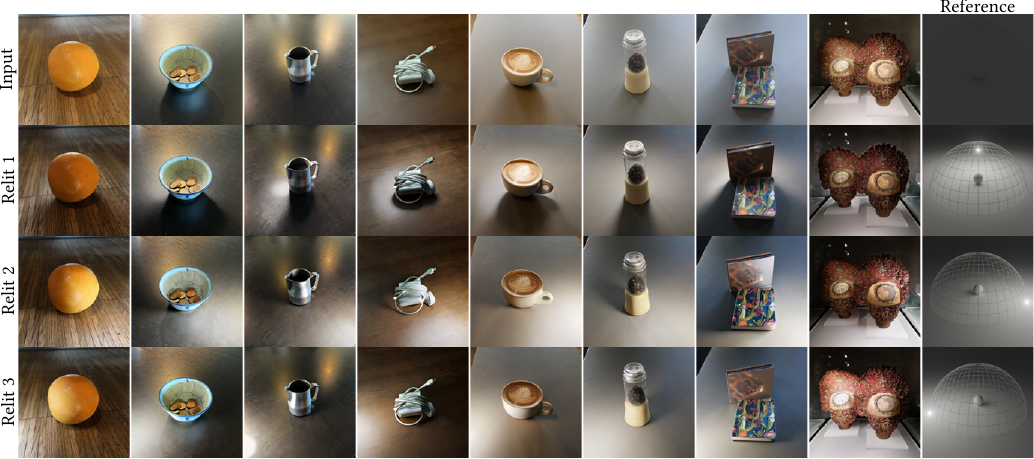}
\caption{\textbf{Object-Level: Qualitative Evaluation}. This result demonstrates generalization to real images, where the model was trained on synthetic dataset: \singleobjecttrain{}. The first row has input images from \datasetRealSingle{} and the last column shows the reference light position. Note: The camera angle is very different for each image and does not exactly match the reference. Please treat the reference as a reference and not absolute. The results show that \moniker{} can insert point lights very close to the object, casting shadows that are plausibly appropriate with respect to the light position. }
\Description{}
\label{fig:in_the_wild_single}
\end{figure*}
\subsection{Object-Level}
\label{eval:object_level} 
In this section, we conduct evaluations on \singleobjecttest{} to evaluate the quality of relighting.
We compare with object-centric baselines that support single-image relighting of a \textit{single object} as input:
(1) WS-SIR \cite{Yi_2023_CVPR} follows a conventional inverse-rendering pipeline based on a Lambertian shading model and represents the light using spherical harmonics. (2) Neural Gaffer \cite{Jin2024NEURIPS_Neural_Gaffer_Relighting} is built on Zero-1-2-3 \cite{liu2023zero1to3}, where the input image is conditioned by an HDRI map represented in both LDR and HDR space (3) IC-Light \cite{iclight} fine-tunes an \emph{image} diffusion model using ControlNet \cite{zhang2023controlnet} and is designed to relight a foreground object by blending an input background image. (4) DiLightNet \cite{zeng2024dilightnet} has a three-step pipeline, where the first stage reconstructs a mesh of the foreground object from depth maps. Further processing is performed in Blender to generate radiance hints that are used as a conditioning signal to relight the input image. (5) Diffusion Renderer \cite{DiffusionRenderer} has two networks, where the neural inverse renderer takes an input video and predicts depth, normal, albedo, metallic and roughness maps. The forward renderer takes these respective maps, and is conditioned on an HDRI map similar to Neural Gaffer to relight.

All of these methods differ significantly by construction, as they represent different paradigms for relighting. For instance, the light representation of IC-Light is a background image, DiLightNet computes radiance hints, WS-SIR requires spherical harmonics, while both Neural Gaffer and Diffusion Renderer use an HDRI map. We render the additional inputs required by each of the methods and provide a detailed explanation in the Supplemental Document.

We report quantitative results in \cref{tab:baseline_comparison}, and show qualitative comparisons in \cref{fig:baseline_fig}. 
WS-SIR predicts noisy surface normals and albedo with baked-in lighting, which leads to incorrect relighting. Neural Gaffer struggles to retain high-frequency details of the foreground object in comparison to other methods. As the method uses an HDRI map, which is an environmental illumination, the relit results have very subtle change in lighting. In contrast, \moniker{} excels in handling local lighting and synthesizes high-frequency lighting effects where the light fall-off is evident. IC-Light struggles to retain the source color of the object after relighting. Moreover, as it is trained to blend foreground and background, the ground texture can act as an additional light source, with background colors bleeding onto the foreground unrealistically. DiLightNet shows good local lighting effects compared to the other baselines, but the foreground often has artifacts that are visible after relighting due to noisy depth estimates. Diffusion Renderer tends to change the identity of the object after relighting, for example, in the last row of \cref{fig:baseline_fig}, the speakers of the radio have the color of the HDRI map (blue sky), possibly because both the forward and inverse renderers are trained separately. Finally, \moniker{} synthesizes shadows of the foreground object onto the background, which is crucial for realism. Our single-image relighting results have plausible shadows and light fall-off that are closely comparable to the path-traced ground-truth versions generated from 3D assets.  

\begin{figure}
    \centering
    \newcommand{\mywidth}{0.7}
    \includegraphics[width=\mywidth\linewidth]{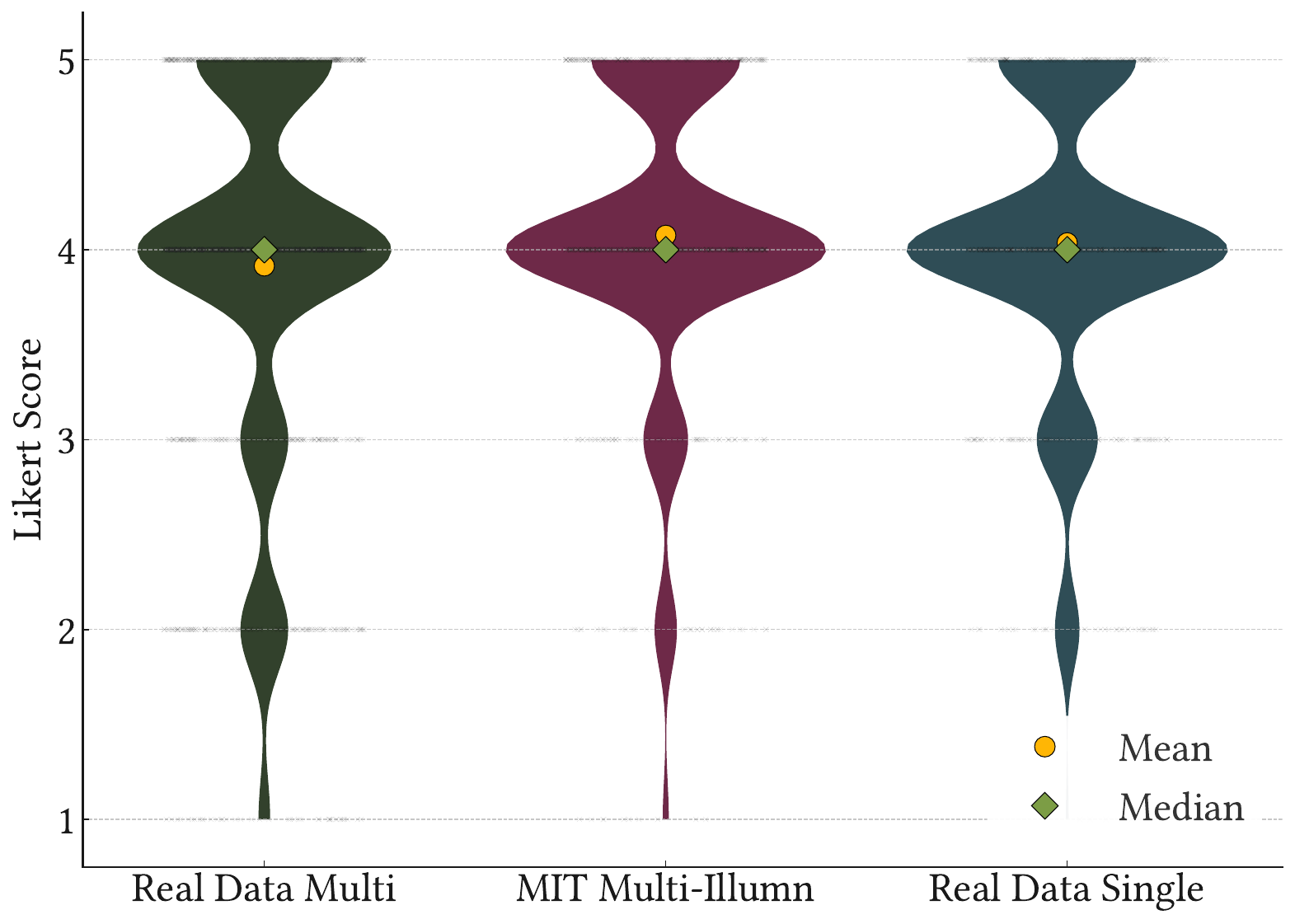} 
    \caption{\textbf{Quantitative Evaluation: Perceptual Study}. Each violin plot shows the distribution of the Likert scores (1 = not realistic, 5 = very realistic) conducted on \datasetRealMulti{}, MIT Multi-Illumination test set and \datasetRealSingle{} respectively. Across all three studies, both the mean and median are close to 4, and the distributions are narrow, indicating that participants consistently perceived the rendered results as ``realistic".}
    \label{tab:user_study}
\end{figure}
\subsubsection{Generalization to Real Images}
\label{sec:in_the_wild_single_object}
In \cref{fig:in_the_wild_single}, we show generalization of \moniker{} on \datasetRealSingle{} captured with a phone (\cref{data_des:test_real}) and the evaluation of the perceptual study in \cref{tab:user_study}. Our results show that \moniker{} can synthesize plausible shadows and reflections that are  appropriate for the materials. For example, in \cref{fig:in_the_wild_single} col 1, the orange retains its texture properties after relighting, which is necessary for realism. Similarly, from row 2 and 3, the pitcher and the coins appear metallic, indicating that the model is able to recognize and model different materials. Furthermore, row 2 shows that our method removes most of the source shadow on the ground such that it is indicative of the light position, making the predictions look convincing. In row 3, since the light is above and slightly behind, there is a strong shadow cast in front of the object. These results show that \moniker{} can synthesize local illumination effects from a single-image, without explicit information about geometry or material assets.

\begin{figure*}
    \centering
    \newcommand{\mywidth}{1}
    \includegraphics[width=\mywidth\textwidth]{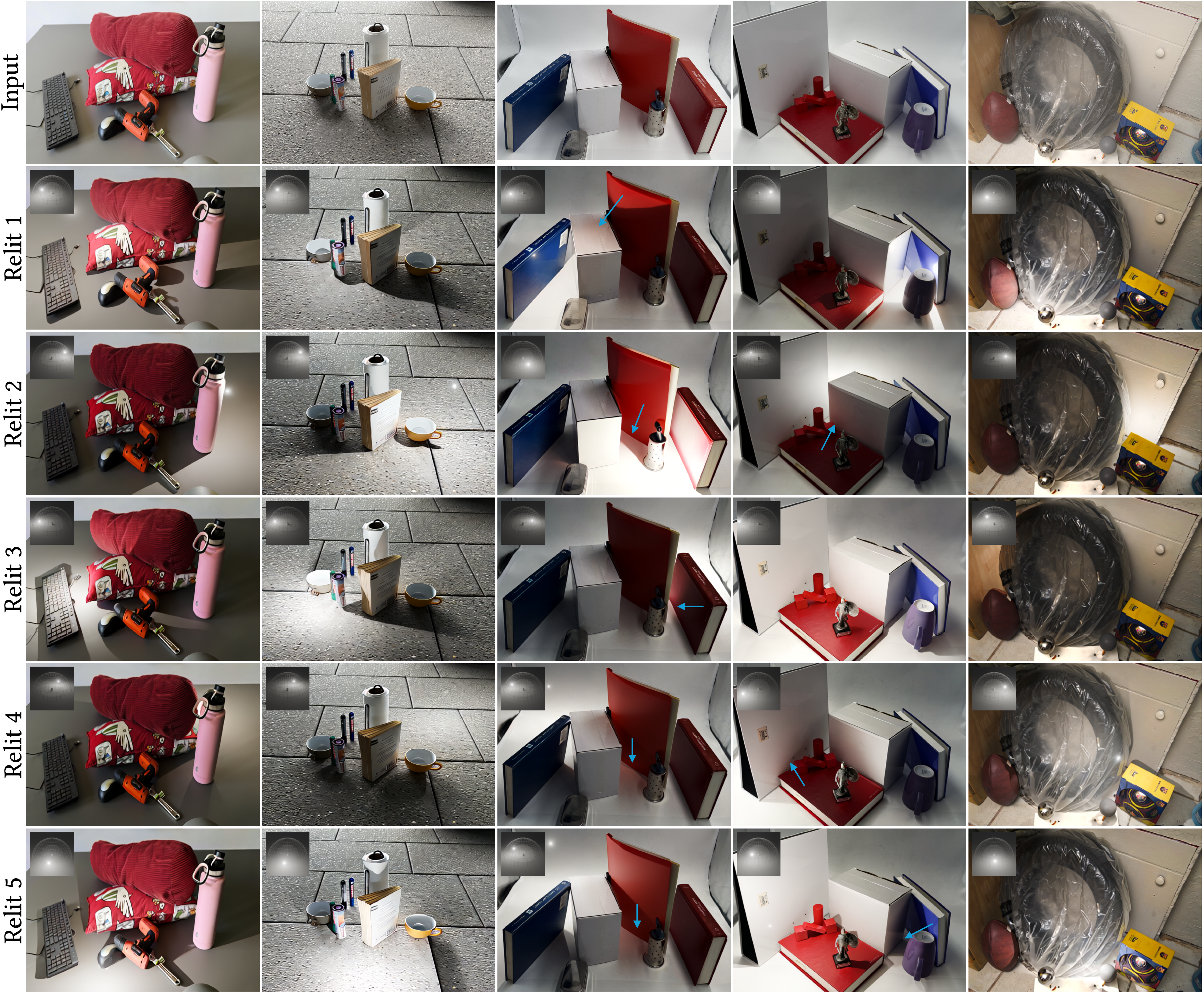}
    \caption{\textbf{Qualitative evaluations: Scene-Level:} These results demonstrate generalization to real images, where the model was trained on a synthetic dataset: \multiobjecttrain{}. The first four columns contain images captured on the phone (\datasetRealMulti{}) and the last column is from the test set of the MIT Multi-Illumination Dataset \cite{murmann19}. The results show that \moniker{} can synthesize indirect lighting effects such as diffuse interreflections resulting in color bleeding (see blue arrows) and convincing shadows that look plausible with respect to the position of the light. }
    \Description{}
    \label{fig:in_the_wild_multi}
\end{figure*}

\subsection{Scene-Level}
\label{sec:scene_level}
In this section, we evaluate the relighting quality on complex scenes containing multiple objects.  
\subsubsection{Generalization to Real Images}
In \cref{fig:in_the_wild_multi}, we show generalization of \moniker{} on \datasetRealMulti{} (\cref{data_des:test_real}) and on the test set of the MIT Multi-Illumination \cite{murmann19} dataset. The evaluation of the perceptual study is reported in \cref{tab:user_study}. From the results in \cref{fig:in_the_wild_multi}, col 1, we observe the shadow of the water bottle changing with the direction of the inserted point light. Similarly, from col 2, we see that the book casts a strong shadow on the mugs placed in both row 2 and row 3, as the position of the light changes. From the qualitative results (col 3 and 4), we  observe diffuse inter-reflections when the light from the inserted point light bounces off the colored object and bleeds onto the ground plane in between. Moreover, \moniker{} can handle materials that are out of the training dataset distribution such as linen pillows and tires wrapped in plastic, retaining the realism of the material after relighting. These results indicate that \moniker{} can handle both direct and indirect lighting effects plausibly. Finally, the perceptual studies reinforce the plausibility of our results after relighting, with participants consistently rating them as realistic.

\subsubsection{Comparisons on MIT Multi-Illumination Dataset}
\begin{table}[htbp]
\centering
\caption{\textbf{Quantitative Evaluation: MIT Multi.-Illumn} comparing against Latent Intrinsics \cite{zhang2024latentintrinsicsemergetraining} and LumiNet \cite{Xing2024luminet} (w/ bypass-decoder). Top: additional metrics (PyTorch impl.). Bottom: Scores taken from Tab. 1 of \cite{zhang2024latentintrinsicsemergetraining, Xing2024luminet} and metrics computed using the authors' implementation (post-discussion) for fairness.}
\resizebox{\linewidth}{!}{%
\begin{tabular}{lcccc}
\toprule
         & RMSE $\downarrow$ & LPIPS $\downarrow$ & SSIM $\uparrow$ & PSNR $\uparrow$ \\
\midrule
Lat. Intr. & 0.1513 & 0.1526 & 0.7574 & 28.454 \\
LumiNet & 0.1862 & 0.1859 & 0.6784 & 28.1718 \\
Ours       & \textbf{0.1129} & \textbf{0.1302} & \textbf{0.7623} & \textbf{28.974} \\
\midrule
SA-AE \cite{Zhongyun2020SA-AE} & 0.317 & - & 0.431 & - \\
S3NET \cite{YangS3Net}         & 0.414 & - & 0.377 & - \\
Lat. Intr                      & 0.222 & - & 0.571 & - \\
LumiNet                       & 0.240 & - & 0.527 & - \\
Ours                           & \textbf{0.219} & - & \textbf{0.584} & - \\
\bottomrule
\end{tabular}
}
\label{tab:baseline_mit}
\end{table}
\begin{figure*}
    \centering
    \newcommand{\mywidth}{1}
    \includegraphics[width=\mywidth\textwidth]{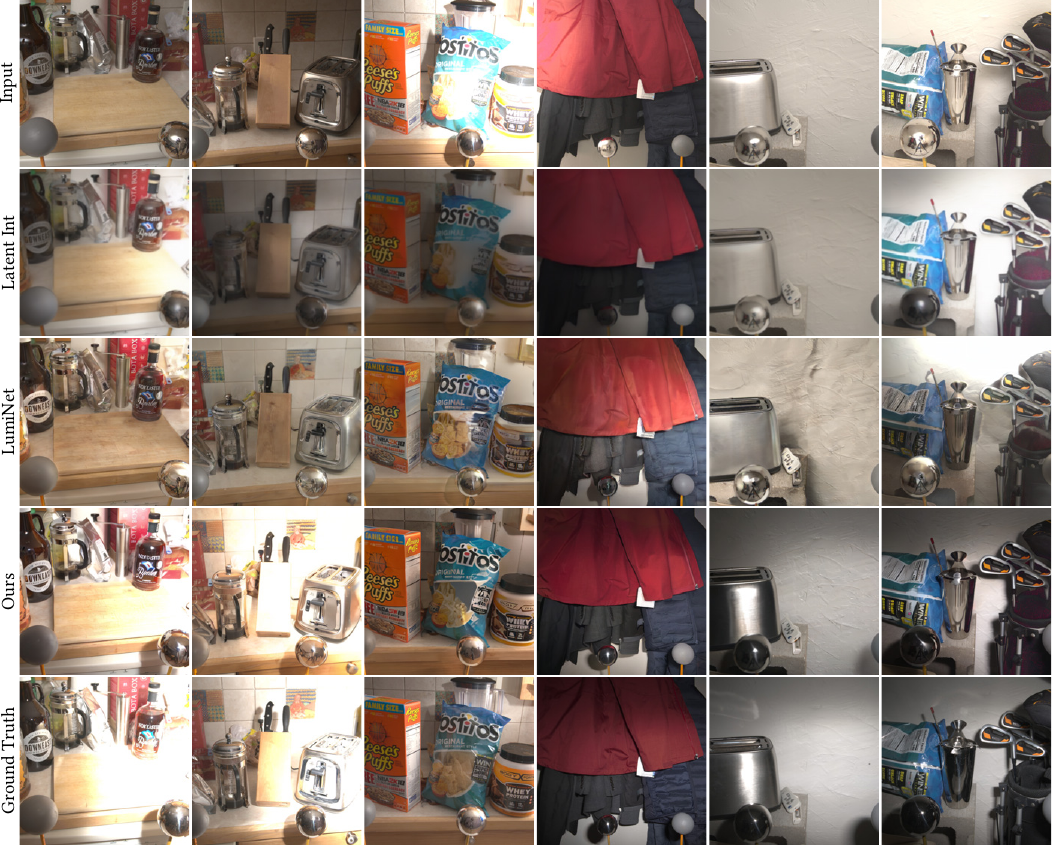}
\caption{\textbf{Scene-Level: Qualitative Evaluation: MIT Dataset - Baseline Comparison} with Latent-Intrinsics~\citep{zhang2024latentintrinsicsemergetraining} and LumiNet~\citep{Xing2024luminet} (w/ bypass-decoder). The results show that \moniker{} generalizes better to complex scenes with multiple cluttered objects than the baselines; our results appear closer to the ground truth. The model was trained on training set of MIT Multi-Illum. \cite{murmann19}.}
\label{fig:mit_comparison_fig}
\end{figure*}
\label{sec:mit_evaluations}
We train our method on the train split of the MIT Multi-Illumination Dataset~\cite{murmann19}, which contains a single mounted light source rotated in 25 lighting directions across 945 scenes. We adapt it to our setting by creating 25 video sequences that have continuous motion (described in the Supplemental Document). We conduct evaluations on the test split (\cref{tab:baseline_mit}) containing 30 scenes and compare with a state-of-the-art encoder-based method, Latent Intrinsics \cite{zhang2024latentintrinsicsemergetraining} and a method build on Stable Diffusion \cite{rombach2021highresolution} LumiNet~\cite{Xing2024luminet} (latest version w/ bypass-decoder). From \cref{fig:mit_comparison_fig}, we see that \moniker{} preserves scene details and synthesizes very realistic relighting effects compared to Latent Intrinsics and LumiNet. Primarily, from col 1 we see that, when the light is very close to the camera, our prediction contains a bright spot, similar to the ground-truth where the input image has an ambient light source. Furthermore, col 3 has an overexposed input image, but \moniker{} can still hallucinate the chips packet (empty), while latent intrinsics has a white artifact in the prediction. While LumiNet retains the scene well, the model struggles to transfer the target lighting as desired and its predictions show only subtle changes in lighting. Finally, in col 5, our result has a sharp specularity on the toaster that is not present in the input and is similar to the ground-truth specularity. 
\section{Limitations and Future Work}
\begin{figure}
    \centering
    \newcommand{\mywidth}{0.45}
\includegraphics[width=\mywidth\textwidth]{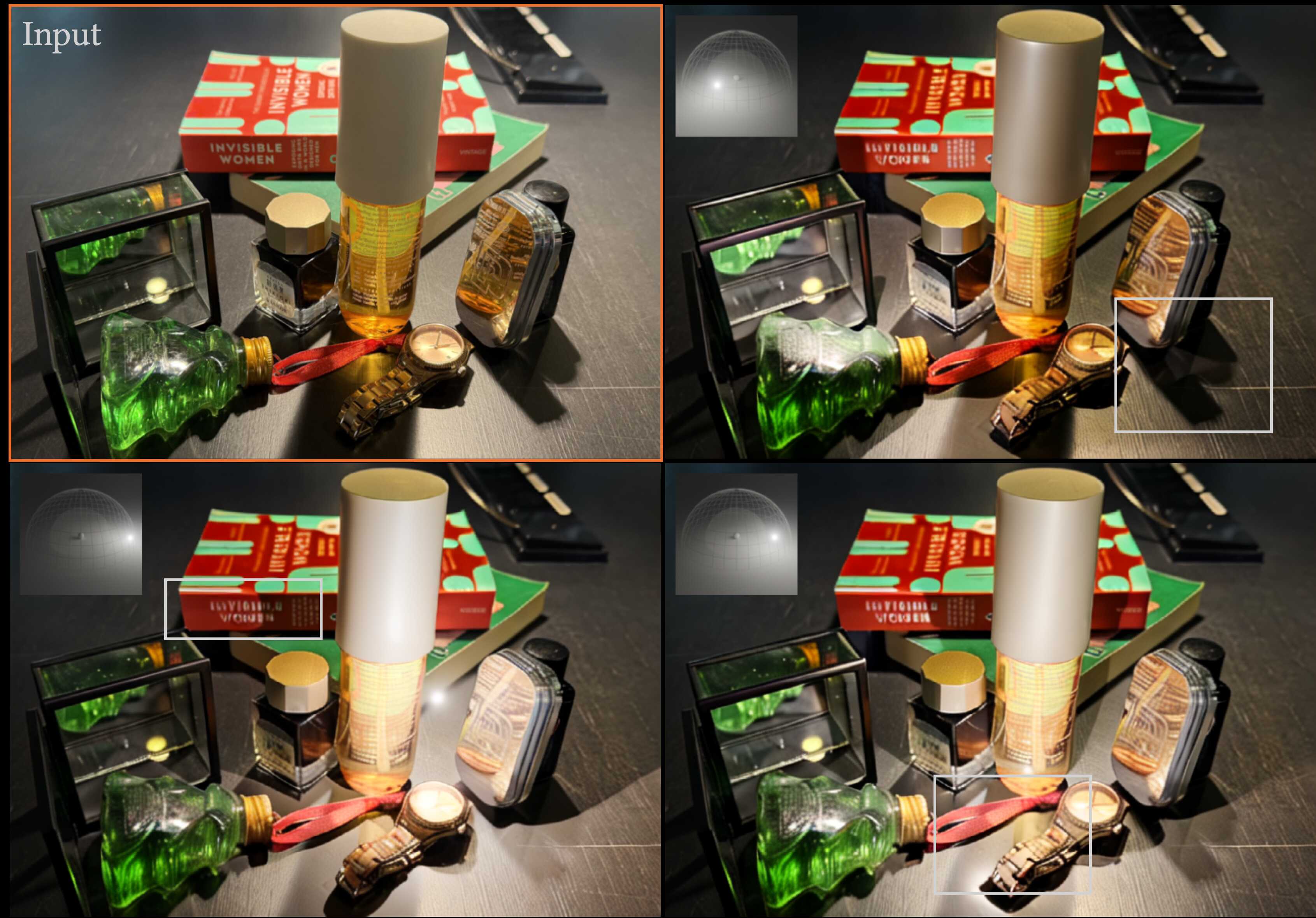}
    \caption{\textbf{Limitations}: Our model is unable to handle complex effects like caustics or complex transmission through liquids
    (\eg the transparent plastic bottle with liquid). 
    Fine-details (\eg book title) are not retained and this limitation is inherited from the stable diffusion VAE. Strong shadows are baked-in as our method focuses on adding in an additional light and not on removing the existing light. }
    \Description{}
    \label{fig:limitations}
\end{figure}

First, our work focuses on a single point light source. There is potential and scope for future research to extend this approach to an HDRI map to fully alter the ambient environment of the scene. Furthermore, in some of the cases we observe that there are shadows that are not removed and baked-in as shown in \cref{fig:limitations}. Since our method focuses on adding in an additional light source and not on removing the existing light, this extends beyond the intended scope of this work. We hypothesize that a training dataset with scenes containing dominant shadows may help the model to learn to remove them. Similarly, our model currently is unable to handle caustics or sub-surface scattering (SSS) as  our synthetic dataset does not contain objects exhibiting such properties. Caustics and SSS are non-trivial to model and the community has dedicated research to tackle its complexity. Training on an OLAT dataset could help in modeling SSS. Finally, stable video diffusion relies on a VAE encoder-decoder and that limits its ability to reproduce fine details, which is a drawback when precision is essential. Additionally, the model tends to hallucinate incorrect specularities and shadows that may appear convincing, but may be physically incorrect. Our  method also inherits the slow inference speeds typical of diffusion models, hindering its use for real-time rendering. However, there is significant progress in the video generation field where powerful models are emerging that are both faster and have better quality results, thus, there is a lot of potential in this direction. 

\section{Conclusion}
In this work, we focus on single-image relighting, with a point-light source, to observe both direct and indirect lighting effects. We reformulate the static relighting task as a dynamic one, where the light is in motion. To model this change, we leverage a video diffusion model and control the intensity and position of the point light by enhancing the video model with control signals. Our results demonstrate generalization to real-world captured scenes with \emph{unseen} objects where \moniker{} convincingly synthesizes shape- and light-appropriate shadows including complex indirect lighting effects such as color bleeding. Moreover, our experiments on multi-object scenes indicate that \moniker{} has an implicit representation of the objects in the scene as it retains the material properties of unseen materials for real-images even though it is fine-tuned on only synthetic data. Our experiments demonstrate that fine-tuning video diffusion models is a promising alternative to traditional, explicit image-relighting pipelines.

\begin{acks}
We thank Peter Kulits for discussions, proofreading, Suraj Bhor for helping us with the baselines, and Nikos Athanasiou for providing feedback. We are very grateful for the support provided by Tsvetelina Alexiadis, Florentin Doll, Markus Höschle, Arina Kuznetcova, Tomasz Niewiadomski, Taylor Obersat and Tithi Rakshit for help with dataset curation and conducting Perceptual Studies. We thank Rick Akkerman for his help with code and Prerana Achar, Radek Daněček, Shashank Tripathi and Anastasios Yiannakidis for their support with visualizations. Finally, we thank Pramod Rao for fruitful discussions and innumerable support. \\Disclosure: While MJB is a co-founder and Chief Scientist at Meshcapade, his research in this project was performed solely at, and funded solely by, the Max Planck Society.
\end{acks}

\bibliographystyle{ACM-Reference-Format}
\bibliography{main}
\balance

\clearpage
\appendix
\section{Ablation Study}
\subsection{Analysis on the 5D Vector}
As described in Section 3.2 of the main paper, the control signal of our model is a sequence of 5D vectors $[\Vec{l}_1, \dots, \Vec{l}_N]$, where each $\Vec{l}i = [ (\phi_i, \theta_i, r_i, I{p_i}, I{e_i}); i=1\dots N ]$. Here, $I{p_i}$ and $I_{e_i}$ are scalar values representing the intensity of the point light and environment light, respectively. Both values are normalized such that $I_{e_i}, I_{p_i} \in [0, 1]$.

\begin{figure}
    \centering
    \newcommand{\mywidth}{0.45}
\includegraphics[width=\mywidth\textwidth]{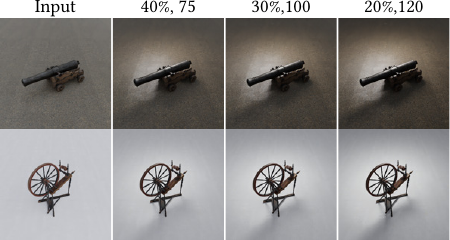}
\caption{\textbf{Ablation Study: Examples of \singleobjecttrain{} scenes}. The lighting condition for $I_{e_i}$ is in percentage and $I_{p_i}$ in lumens (Blender). }
\Description{}
\label{fig:sensitivity_analysis_fig}
\end{figure}

We create additional synthetic datasets under the \singleobjecttrain{} setting, using the same objects and setup described in Section 3.3 of the main paper. As shown in \cref{fig:sensitivity_analysis_fig}, $I_{e_i}$ is reduced from $1.0 \xrightarrow{} [0.4, 0.3, 0.2]$, while $I_{p_i}$ is increased from $0.0 \xrightarrow{} [75, 100, 120]$ lumens. For evaluation, we generate a test set using the \singleobjecttest{} configuration and render 8 light trajectories for each object under the three different intensity pairs, resulting in 1.8k videos.

\begin{figure}
    \centering
    \newcommand{\mywidth}{0.45}
\includegraphics[width=\mywidth\textwidth]{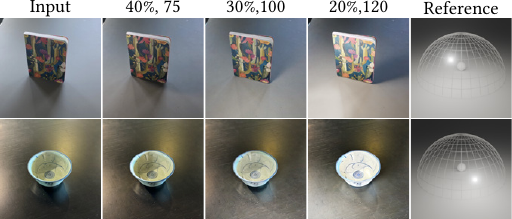}
\caption{\textbf{Ablation Study: Generalization to real images}. The first column has the input image captured on a phone and the last column has the reference light position. The results are for three different intensity values where $I_{e_i}$ is in percentage and $I_{p_i}$ in lumens (Blender). The model trained with ($I_{e_i}=20\%$, $I_{p_i}=120$ lumens) has better results where the source light is sufficiently dimmed such that the point light acts as the dominant light source as intended. }
\Description{}
\label{fig:sensitivity_analysis_results}
\end{figure}

Quantitative results in \cref{tab:ablation_5d} show that the setting ($I_{e_i}=20\%$, $I_{p_i}=120$ lumens) achieves the best score. Qualitatively, as shown in \cref{fig:sensitivity_analysis_results}, this setting also generalizes better to real images. With ($40\%, 75$), the source light is not dimmed enough and the inserted point light is barely visible. However, with ($30\%, 100$), the point light is visible but its intensity is subtle and the source light remains as the dominant light source. In contrast, ($20\%, 120$) produces clear shadows because the inserted point light acts as the dominant light source as intended and the source light is sufficiently dimmed. 

\begin{table}[htbp]
\centering
\caption{\textbf{Ablation Study: Intensity Values}. $I_{e_i}$ is in percentage and $I_{p_i}$ is given by lumens (Blender). }
\begin{tabular}{lccccc}
\toprule
$I_{e_i}$ & $I_{p_i}$ & RMSE $\downarrow$ & LPIPS $\downarrow$ & SSIM $\uparrow$ & PSNR $\uparrow$ \\
\midrule
40\% & 75 & 0.0378 & 0.0273 & 0.9687 & 40.1602\\
30\% & 100 & 0.0353 & 0.0243 & 0.9722 & 40.0659 \\
20\% & 120 &\textbf{0.0330} & \textbf{0.0207} & \textbf{0.9771} & \textbf{40.5039}  \\
\bottomrule
\end{tabular}
\label{tab:ablation_5d}
\end{table}

Moreover, the synthetic dataset is built with 620 HDRI maps (described in main paper, Section 3.3), each with a random rotation applied to the environment. Hence, the dataset has a wide range of ambient lighting, from bright to quite dull and, as seen in \cref{fig:sensitivity_analysis_fig}, ($20\%, 120$), our chosen setting strikes a good balance between both. Moreover, real-world images have complex lighting, and these results indicate that training with this synthetic setup helps with better generalization.

\begin{figure}
\centering
    \newcommand{\mywidth}{0.45}
\includegraphics[width=\mywidth\textwidth]{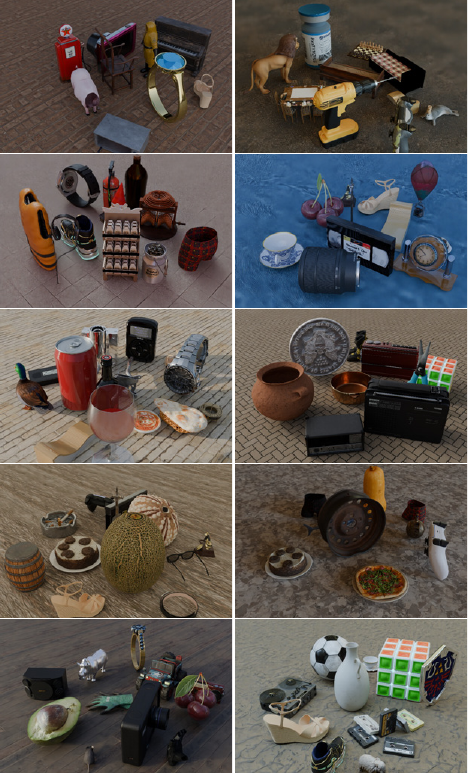}
    \caption{Examples of \multiobjecttrain{} scenes.}
    \label{fig:multiobject}
\end{figure}
\section{\datasetname}
\label{sec:dimming}
\begin{figure}
\centering
    \newcommand{\mywidth}{0.45}
\includegraphics[width=\mywidth\textwidth]{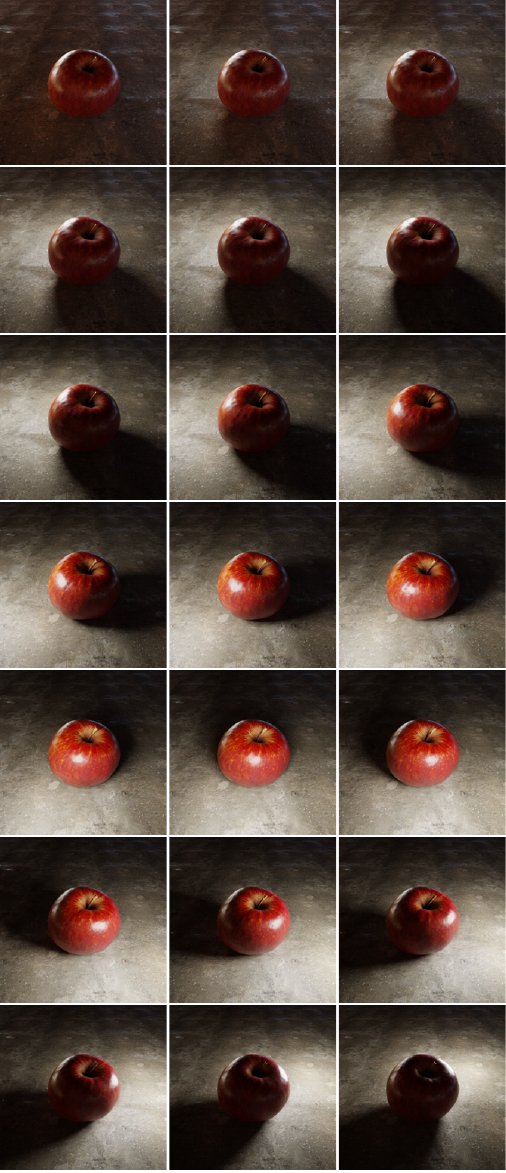}
    \caption{\textbf{Object lighting.} In the first six frames, the HDRI lighting gradually dims while a point light fades in. In the subsequent frames, the point light follows a trajectory, moving from the left side of the object to the bottom, and then to the right.}
    \label{fig:object_lighting}
\end{figure}

As mentioned in the main paper, the point light at each frame is represented by a 5-D vector. For the \singleobjecttrain{} scenario, we create trajectories of 14 frames via linear interpolation of the polar coordinates. During the first four frames, the intensity of the environment map is dimmed from 1.0 to 0.2. Meanwhile, the intensity of the point light is increased from 0.0 at frame 1, to 120 lumens making it visible from frame 2. During the dimming process the position of the point light remains static; the light movement occurs between frames 5 and 14. For the \multiobjecttrain{} setup, we use the same dimming process but create longer trajectories of 25 frames, and the light movement occurs between frames 7 and 25. 
A rendered example of the dimming and movement can be seen in \cref{fig:object_lighting}.

\begin{figure}
\centering
    \newcommand{\mywidth}{0.45}
\includegraphics[width=\mywidth\textwidth]{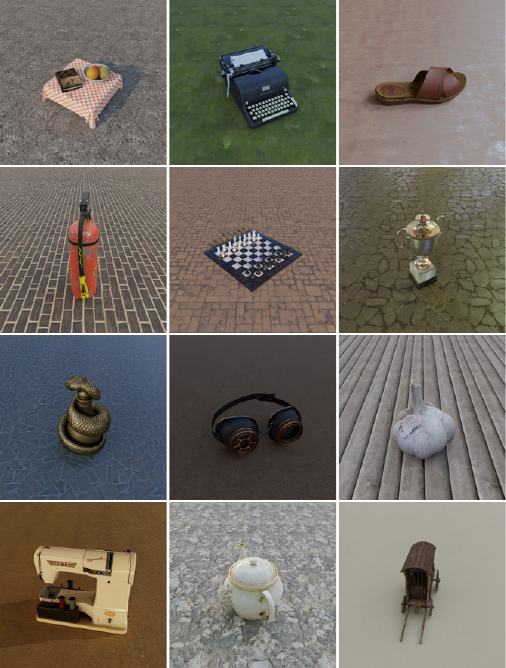}
    \caption{Examples of \singleobjecttrain{} scenes. }
    \label{fig:singleobject}
\end{figure}
With regards to scene construction, examples of \singleobjecttrain{} scenes can be seen in \cref{fig:singleobject}, while \multiobjecttrain{} scenes are shown in \cref{fig:multiobject}.

\paragraph{Rendering} We create our dataset using Blender 3.2.2 and render the scene using the Cycles path-tracing engine with denoising enabled. For the \singleobjecttrain{} and \multiobjecttrain{} scenarios, the number of samples-per-pixel is set to 1024 and 512 respectively, while the number of ray bounces for diffuse, glossiness, transparency and transmission is set to 8 in both setups.

\begin{figure}[t]
    \centering
    \includegraphics[height=1.42in]{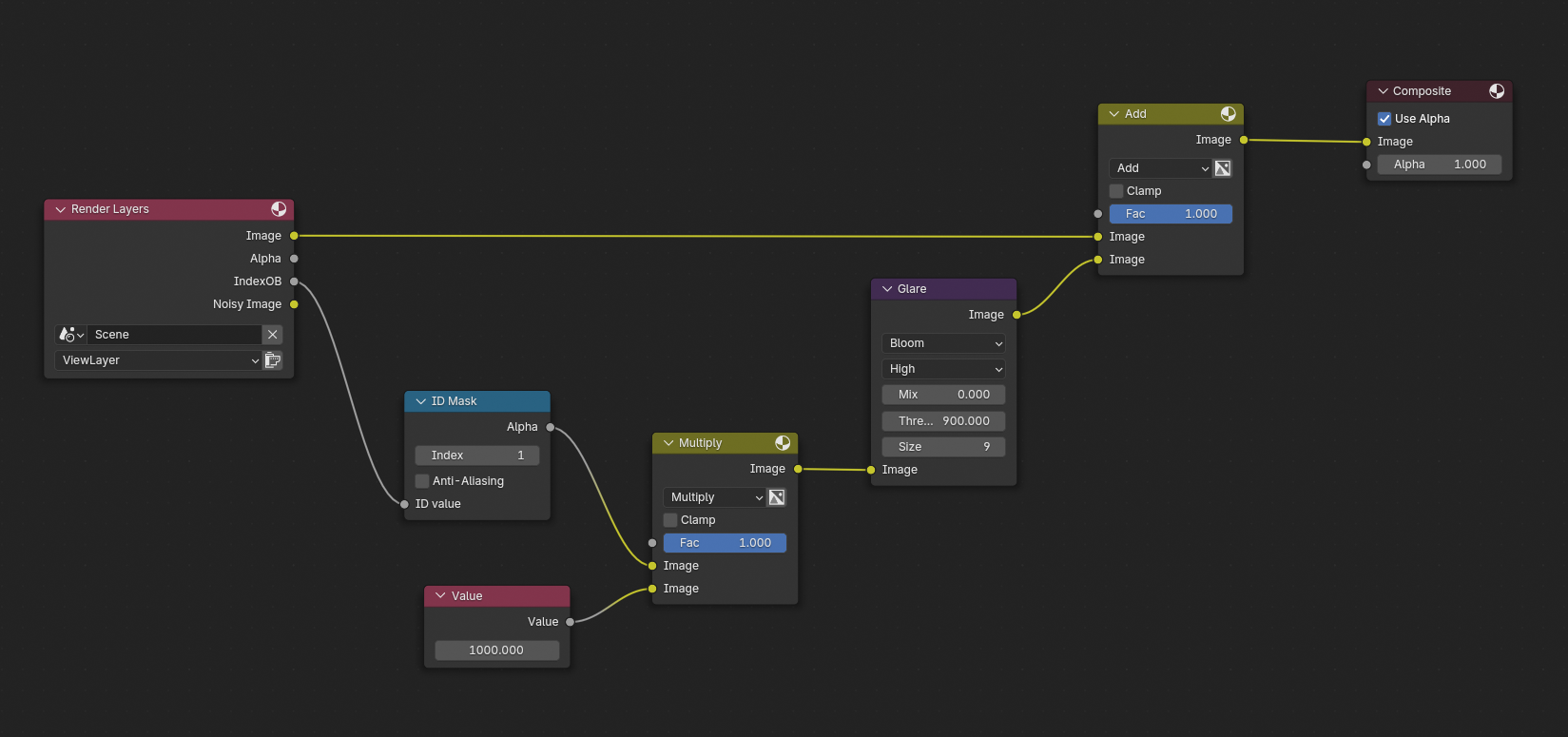}
    \caption{\textbf{Post-processing.} The node tree used in the Blender compositor to achieve the glow effect.}
    \label{fig:compositing}
\end{figure}

\paragraph{Post-processing. } For the multi-object portion of the dataset, we enhance the appearance of the moving point light by adding a glow effect using the Blender compositor for the light to appear \textit{fairy-like}. This effect is achieved by: (a) placing a small sphere at the light's position and rendering a mask of the sphere; and (b) refining the mask and applying a \texttt{Glare} node with the glare type set to \texttt{Bloom}. Figure~\ref{fig:compositing} illustrates the node tree used for this post-processing step.

\begin{figure}
    \centering
    \newcommand{\mywidth}{0.45}
\includegraphics[width=\mywidth\textwidth]{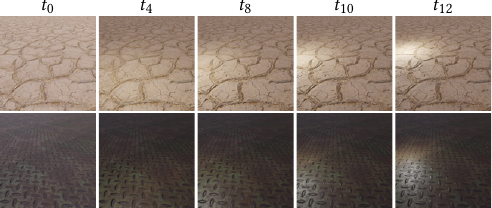}
    \caption{\textbf{Implementation Details.} Background images provided to IC-Light visualized at different frames. The generated background images match the lighting of each video.}
    \label{fig:ic_light_condition}
\end{figure}

\section{Training details. }
We fine-tune our model based on Stable Video Diffusion \cite{blattmann2023svd}. We use a learning rate of $5 \times 10^{-5}$ and the training is conducted on 8 A100 gpus with a batch size of 16. We finetune using the AdamW optimizer for 20,000 iterations. The \singleobjecttrain{} is trained on a resolution of 512x512 and \multiobjecttrain{} on 640x448. 

\begin{figure*}
    \centering
    \newcommand{\mywidth}{1}
\includegraphics[width=\mywidth\textwidth]{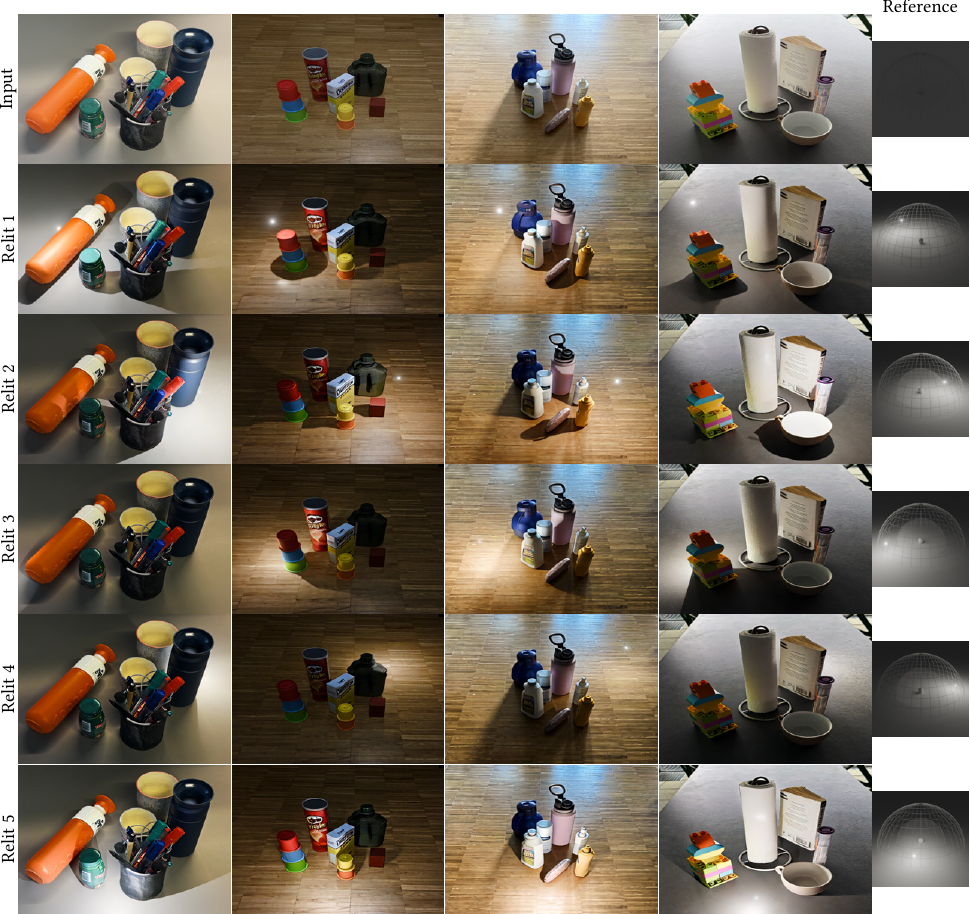}
    \caption{\textbf{Qualitative evaluations: Scene-Level:} The first row has the input images of real-world objects captured on the phone and the last column has the reference light position. This result demonstrates generalization to real images, where the model was trained on a synthetic dataset: \multiobjecttrain{}. The results indicate that \moniker{} can insert a point light and synthesize convincing shadows that look plausible with respect to the position of the light. Additionally, shadows of one object are also cast on the other object(s). }
    \Description{}
    \label{fig:in_the_wild_multi_add_1}
\end{figure*}
\section{Evaluation}
\subsection{Perceptual Study - Details}
The study consists of 5 catch trials for each batch of 8 videos. If a participant failed 3 or more catch trials, their evaluation was removed. In total, after removing subjects who failed the catch trials, there were 48 participants for the study. \datasetRealSingle{} and \datasetRealMulti{} had 400 videos each, while the MIT Multi Illumination testset images had 300 videos. Furthermore, each participant was given 4 videos initially as a warm-up to get acclimatized to the study. The results from the warm-up were are excluded from the final results. 

\begin{figure*}
    \centering
    \newcommand{\mywidth}{1}
\includegraphics[width=\mywidth\textwidth]{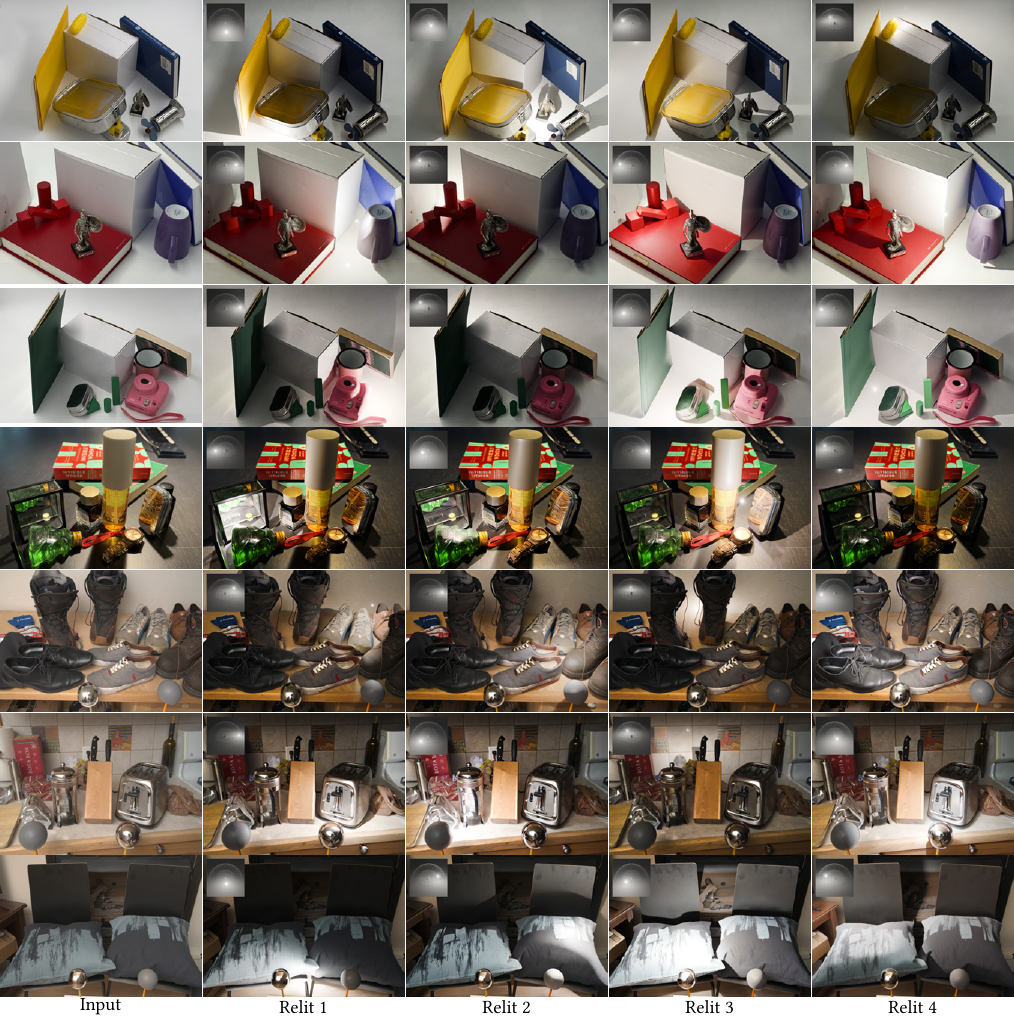}
    \caption{\textbf{Qualitative evaluations: Scene-Level:} This result demonstrates generalization to real images, where the model was trained on a synthetic dataset: \multiobjecttrain{}. The first four rows contain real-world images captured on phone and the following three rows are images from the test set of the MIT Multi-Illumination Dataset \cite{murmann19}. The results show that \moniker{} can synthesize indirect lighting effects such as diffuse interreflections resulting in color bleeding and convincing shadows that look plausible with respect to the position of the light. }
    \label{fig:in_the_wild_multi_add_2}
\end{figure*}
\subsection{Baselines - Implementation Details}
In the main paper we compare against baselines that represent different paradigms: IC-Light \cite{iclight}, DiLightNet \cite{zeng2024dilightnet}, Neural Gaffer \cite{Jin2024NEURIPS_Neural_Gaffer_Relighting}, Diffusion Renderer \cite{DiffusionRenderer} and an inverse rendering baseline WS-SIR \cite{Yi_2023_CVPR} (a reconstruction+rendering baseline). We adapt them to our setting as follows:
For IC-Light, we render the background image of each test frame for \singleobjecttrain{} to exactly match the lighting of the ground truth relit image. We follow the dimming process and the movement of the point light as described in \cref{sec:dimming}. The input to the IC-Light network for each video is then the rendered background images for 14 frames, together with the input image (that is the original test image) along with the object mask. The background images are 768x768 pixels and we upsample the input image as IC-Light takes high resolution inputs. 
We render HDRI maps for each frame of the video to match the lighting of the ground truth. We use a panoramic camera in Blender and create a sphere with an emissive material to represent the point light. For WS-SIR, Neural Gaffer, DelightNet and Diffusion Renderer we provide the HDRI map and use the authors implementation accordingly. Further, we provide object mask to segment the albedo and normal maps prediction.  

\begin{table}[h]
\centering
\caption{We provide the indices of the light directions of the MIT dataset that are used to generate 25 different trajectories that are continous. We ensure each light direction is fed as the input. }
\label{tab:mit_trajectories}
\resizebox{\columnwidth}{!}{%
\begin{tabular}{@{}llcc@{}}
\toprule
\textbf{Trajectories} & \textbf{Indices} \\
\midrule
1 & [23, 11, 0, 10, 1, 17, 6, 15, 5, 13, 12, 4, 16, 14] \\
2 & [14, 12, 4, 16, 15, 5, 13,  7, 11, 0, 10, 1, 17, 18] \\
3 & [0, 11, 23, 24, 2, 22, 3, 19, 18, 17, 9, 8, 12, 13] \\
4 & [11, 23, 24, 2, 22, 3, 19, 18, 17, 9, 8, 12, 13, 5] \\
5 & [23, 24, 2, 22, 3, 19, 18, 17, 9, 8, 12, 13, 5, 15] \\
6 & [24, 2, 22, 3, 19, 18, 17, 9, 8, 12, 13, 5, 15, 16] \\
7 & [2, 22, 3, 19, 18, 17, 9, 8, 12, 13, 5, 15, 16, 4] \\
8 & [12, 4, 16, 15, 5, 13,  7, 11, 0, 10, 1, 17, 18, 19] \\
9 & [4, 16, 15, 5, 13,  7, 11, 0, 10, 1, 17, 18, 19, 3] \\
10 & [16, 15, 5, 13,  7, 11, 0, 10, 1, 17, 18, 19, 3, 22] \\
11 & [5, 13,  7, 11, 0, 10, 1, 17, 18, 19, 3, 22, 2, 24] \\
12 & [1, 10, 0, 11, 23, 24, 2, 22, 3, 19, 18, 17, 9, 8] \\
13 & [3, 19, 18, 17, 9, 8, 11, 7, 13, 5, 15, 16, 4, 14] \\
14 & [6, 17, 1, 10, 0, 11, 23, 24, 2, 22, 20, 15, 16, 4] \\
15 & [7, 11, 0, 10, 1, 17, 18, 19, 3, 22, 21, 13, 12, 4] \\
16 & [8, 9, 17, 18, 19, 3, 22, 2, 24, 23, 11, 0, 10, 1] \\
17 & [9, 8, 11, 23, 24, 2, 22, 3, 19, 18, 17, 9, 8, 12] \\
18 & [10, 0, 11, 23, 24, 2, 22, 20, 15, 16, 4, 12, 13, 5] \\
19 & [13, 5, 15, 16, 4, 12, 7, 24, 2, 22, 3, 19, 18, 17] \\
20 & [17,1, 10, 0, 11, 23, 24, 2, 22, 20, 15, 16, 4, 12] \\
21 & [19, 20, 21, 24, 23, 11, 0, 10, 9, 16, 15, 5, 13, 12] \\
22 & [20, 21, 24, 23, 11, 0, 10, 9, 16, 15, 5, 13, 12, 4] \\
23 & [21, 24, 23, 11, 0, 10, 9, 16, 15, 5, 13, 12, 4, 14] \\
24 & [22, 2, 24, 23, 11, 0, 10, 1, 18, 6, 15, 5, 13, 12] \\
25 & [18, 19, 20, 21, 24, 23, 11, 0, 10, 9, 16, 15, 5, 13] \\

\bottomrule
\end{tabular}}
\end{table}

\subsection{Additional Results}
Additional results of in-the-wild generalization where the model is trained on \multiobjecttrain{} are shown in \cref{fig:in_the_wild_multi_add_1} and \cref{fig:in_the_wild_multi_add_2}.

\section{MIT Multi-Illumination Dataset}
The MIT Multi-Illumination Dataset captures indoor scenes of the real world and has cluttered real-world objects composed of different materials. There are 945 scenes in the training set captured in different rooms such as the living room, kitchen, bathroom, etc. The test set has 30 scenes captured in different rooms, and the objects that appear in the test set are not part of the training set. Every scene is captured under 25 fixed lighting conditions, where the mounted light is rotated and the scene is lit by the light that bounces off the walls and ceilings. To adapt it to our setting, we consider each lighting condition as an input image and construct 25 trajectories where each trajectory has 14 light directions that are selected to form a continous motion of light. We provide the order of the indices of each trajectory in \cref{tab:mit_trajectories}. We downsample the images to size 512$\times$768. However, our quantitative comparisons against Latent-Intrinsics \cite{zhang2024latentintrinsicsemergetraining} and LumiNet \cite{Xing2024luminet} in the main paper (Tab 3) is performed on the center cropped region of 512x512 for fairness.

\end{document}